\documentclass{article}

    \PassOptionsToPackage{numbers, compress}{natbib}

\usepackage[final]{neurips_2021}

\usepackage[utf8]{inputenc} 
\usepackage[T1]{fontenc}    
\usepackage{hyperref}       
\usepackage{url}            
\usepackage{booktabs}       
\usepackage{amsfonts}       
\usepackage{nicefrac}       
\usepackage{microtype}      
\usepackage{xcolor}         
\usepackage{amsmath}
\newcommand{\norm}[1]{\left\lVert#1\right\rVert}
\usepackage{amssymb}
\usepackage{graphicx}
\usepackage{caption}
\usepackage{float}
\usepackage{epsfig}
\usepackage{capt-of}
\usepackage{multirow}

\definecolor{MyPurple}{rgb}{0.6,0.25,0.8}
\definecolor{MyBlue}{rgb}{.11, .63, .95}
\definecolor{MyBlue2}{rgb}{.18,.2,.98}
\definecolor{MyRed}{rgb}{0.7,0.02,0.02}
\definecolor{MyGreen}{rgb}{0.4,0.78,0.3}
\definecolor{MyOrange}{rgb}{0.94,0.39,0.19}

\setcitestyle{square}

\title{\emph{GAN Mask R-CNN:} Instance Semantic Segmentation Benefits from Generative Adversarial Networks}
\title{\emph{Mask R-CNN} as a Generative Adversarial Network}
\title{Thinking Instance Semantic Segmentation as a Generative Adversarial Network}
\title{Instance Semantic Segmentation Benefits from Generative Adversarial Networks}

\author{%
  Quang H. Le \\
  Massachusetts Institute of Technology\\
  Cambridge, MA 02139 \\
  \texttt{quangle@mit.edu} \\
   \And
   Kamal Youcef-Toumi \\
   Massachusetts Institute of Technology \\
   Cambridge, MA 02139 \\
   \texttt{youcef@mit.edu} \\
   \AND
   Dzmitry Tsetserukou \\
   Skoltech \\
   Moscow, Russia, 121205 \\
   \texttt{D.Tsetserukou@skoltech.ru} \\
   \And
   Ali Jahanian \\
   Massachusetts Institute of Technology \\
   Cambridge, MA 02139 \\
   \texttt{jahanian@mit.edu} \\
}

\begin{document}

\maketitle
\begin{abstract}
In design of instance segmentation networks that reconstruct masks, segmentation is often taken as its literal definition -- assigning each pixel a label. This has led to thinking the problem as a template matching one with the goal of minimizing the  loss between the reconstructed and the ground truth pixels. Rethinking reconstruction networks as a generator, we define the problem of predicting masks as a GANs game framework: A segmentation network generates the masks, and a discriminator network decides on the quality of the masks. To demonstrate this game, we show effective modifications on the general segmentation framework in Mask R-CNN. We find that playing the game in feature space is more effective than the pixel space leading to stable training between the discriminator and the generator, predicting object coordinates should be replaced by predicting contextual regions for objects, and overall the adversarial loss helps the performance and removes the need for any custom settings per different data domain. We test our framework in various domains and report on cellphone recycling, autonomous driving, large-scale object detection, and medical glands. We observe in general GANs yield masks that account for crispier boundaries, clutter, small objects, and details, being in domain of regular shapes or heterogeneous and coalescing shapes. Our code for reproducing the results is available publicly.
\end{abstract}

\begin{figure}[!htb]
\begin{minipage}{\textwidth}
	\centering
	\vspace{-.07in}
\includegraphics[width=400pt]{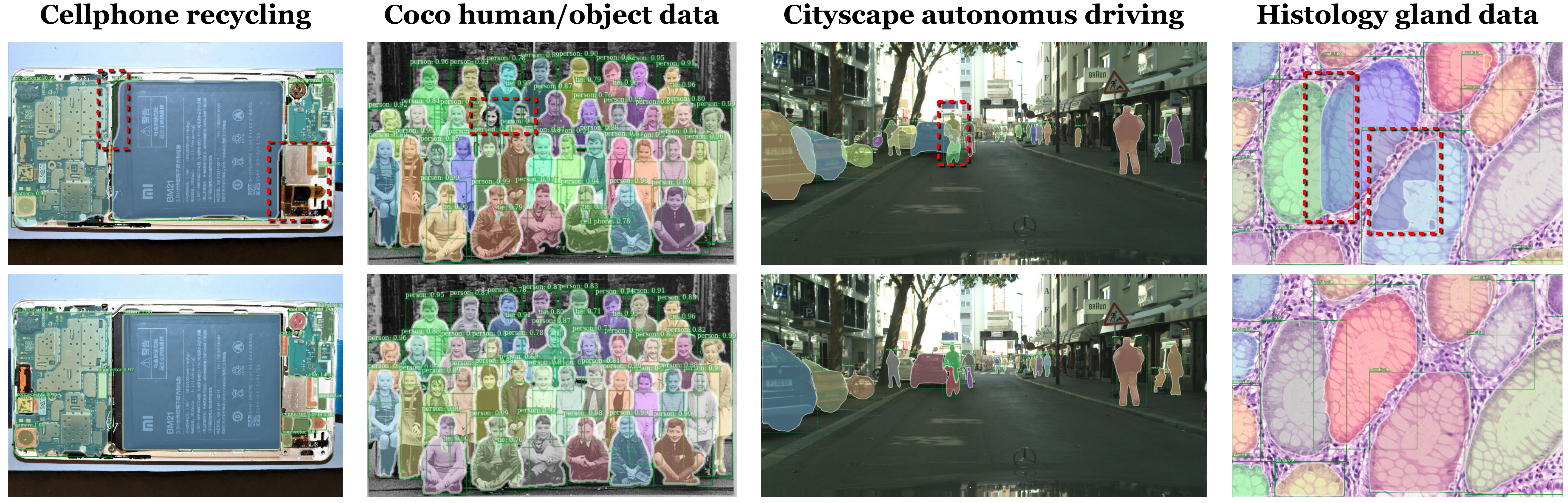}
\vspace{-0.1in}
\captionof{figure}{Instance semantic segmentation has applications in many domains, and each domain may have a specific goal and challenges, e.g., cellphone recycling objects need clear boundaries and seeing small details for disassembling, COCO and Cityscapes are large-scale, and glands are heterogeneous with coalescing pixels. Learning a generic loss in only one generic architecture, and regardless of the kind of underlying domain, we show a GAN loss generally improves the accuracy and quality of the task. Top row: results of the baseline Mask R-CNN. Bottom row: results of incorporating the baseline's loss with a GAN loss. For ease of comparison, we highlight (with the red dotted-boxes) some failures in the baseline results.}
\label{fig:teaser_img}
\end{minipage}
\vspace{0.0in}
\end{figure} 

\section{Introduction}

Inference of instance image segmentation that includes details of crisp boundaries is still a challenging task. While the idea of using generative adversarial networks -- GANs -- \cite{goodfellow2014generative} to produce segmentation mask has been explored, the results are suffering from losses that are based on averaging pixel-wise differences. We suggest to tackle this problem by using a GAN loss merged with Mask R-CNN \cite{he2017mask} and get the best of the both worlds: power of multi-tasking on region-based networks and high-accuracy generations from the GAN loss. One can think of the Mask R-CNN as a mask generator. We can then take these generations and optimize for making them more accurate by using the min-max strategy used in GANs where there is another network -- discriminator -- that tries to distinguish them from the real ground truth masks. We show this strategy outperforms the hand-engineered losses.

Our contribution is three-folded: First, we design a new complementary GAN loss to make Mask R-CNN trainable in the GAN framework. Second, to make this idea work, instead of using images as real samples to discriminator, we use feature maps of the corresponding images. Third, we suggest a way to deploy GAN training on object bounding box detection without directly using only four coordinates, by using Precise RoI Pooling \cite{jiang2018acquisition} to take the computation space to the feature maps surrounded by these coordinates.

Results through extensive experiments show consistent performance gain in both object detection and mask segmentation tasks over the state-of-the-art baseline. Redefining the problem in the GANs framework enables us to use a general loss that is applicable to different data domains and does not need a custom design. For our experiments, we choose various datasets ranging from regular shapes to heterogeneous and coalescing shapes, with different object sizes and details, and different dataset scales: phone recycling \cite{Jahanian_2019_CVPR_Workshops} , Cityscapes autonomous driving \cite{Cordts2016Cityscapes}, COCO objects \cite{lin2014microsoft}, and medical gland dataset \cite{DBLP:journals/corr/Sirinukunwattana16}. Figure~\ref{fig:teaser_img} illustrates some results. Note that without any extra settings, we simply train our model on these datasets and show our design of GAN loss quantitatively and qualitatively improves the baseline model. Our code is available at \url{https://github.com/quangle2110/GAN_Mask-RCNN}.

\section{Related work}

We are interested in designing a generic network that can perform multiple visual tasks in different domains. By multiple visual tasks we mean, object detection and localization, semantic segmentation, instance segmentation and crisp boundaries detection, all together. Many approaches on optimizing ConvNets architectures that can deal with one or several visual tasks have been explored separately and benchmarked on public datasets, and yet the problem is still and active topic of research.
For example, for semantic segmentation, fully convolutional networks, FCN \cite{long2015fully,noh2015learning}, SegNet \cite{badrinarayanan2015segnet} and U-Net \cite{ronneberger2015u} architectures have been successful. Since these networks do not perform well in details and boundaries, several improvements have been suggested \cite{ghiasi2016laplacian,pinheiro2016learning,lin2017refinenet}. 
Here we review the three main components ideas specifically related to our design: FPN, Mask R-CNN, and GANs.

\textbf{FPN:} Driven by the idea that incorporating additional connectivity helps the different types of information to flow across different scales of network depth, recent architectures are based on skip connections from encoder to decoder for using details from different feature maps' level (Laplacian reconstruction in LLR \cite{Girshick_2015_ICCV}, SharpMask \cite{pinheiro2016learning}). One such architecture is FPN \cite{lin2017feature} which demonstrates significant improvement as a generic feature extractor in several visual applications. FPN  utilizes a pyramidal structure (i.e., multi-level) for encoder and decoder, similar to human visual system in multi-scale visual tasks \cite{burt1987laplacian}. FPN further uses an adaptive pooling mechanism for aggregating the feature maps of corresponding encoder/decoder levels together (through lateral skip-connections as well as the inherited layer skip-connections from ResNet \cite{he2016deep}) with losses at each level of the decoder. This architecture gives state-of-the-art for any backbone. FPN is also used as the backbone in Mask R-CNN \cite{he2017mask} which does object detection, localization, and instance semantic segmentation together. In our approach, we extracted the feature map from all the levels of FPN output and send them to GAN discriminator for loss computation.

\textbf{Mask R-CNN} is currently state-of-the-art for object detection and instance segmentation, and part of its strength is due to region-based detection mechanism . In fact this network is an evolution of prior work -- RCNN \cite{Girshick_2014_CVPR}, Fast RCNN \cite{Girshick_2015_ICCV}, and Faster RCNN \cite{ren2015faster}, with refinements in masks, e.g., \cite{liu2018path,chen2019hybrid}. 
The core idea in this family of networks is to scan over the predefined regions called \emph{anchors}. For each anchor, the Region Proposal Network (RPN) does two different type of predictions: the score of it being foreground, and the bounding box regression adjustment to best fit the object. RPN then chooses a fixed number of anchors with highest score, and apply the regression adjustment to get the final proposal for objects prediction at the network head. There are 3 type of prediction branches at the head: bounding box regression, class detection, and mask segmentation. Each head branch extracts the Regions of Interest (RoIs) from the output feature maps of the FPN backbone based on the final proposals from RPN, then feeds them through a combination of fully connected and convolution layers for final predictions. This mechanism works better for us (as we compared to SegNet). Note that there are other instance segmentation works \cite{pinheiro2015learning,dai2016instance} that use the region proposals, however, Mask R-CNN learns all the tasks at the head (i.e., for inference). We elevated Mask R-CNN's training framework by designing a new GAN loss for both masks segmentation and bounding box detection. We use Precise Roi Pooling \cite{jiang2018acquisition} to send the feature map corresponding to the bounding box detection (instead of just 4 coordinates) to GAN discriminator.

\textbf{GAN} losses are applied to many networks and domains, such as \cite{isola2017image,luc2016semantic,xue2018segan}, among which \cite{xue2018segan} is closest to our work, with the exception that we focus on the losses of both the bounding box (which includes incorporating Precise Roi Pooling \cite{jiang2018acquisition}) and mask. Further, unlike their work, our computations are in the feature maps space and thus we send feature maps to GANs' discriminator.

\section{Method}

We propose an adversarial loss design to complement the standard pixel-wise loss used in Mask R-CNN heads. We modify the prediction heads to generate more detail representations of their outcomes, and introduce the corresponding discriminators which, by considering the adversarial loss, can supervise the generative process. In this section, we will first present about the concept of GAN network in general. Then, the detail of the generator - discriminator architecture and objective function implementation are given.

\subsection{Objective}

The learning objective in training GANs \cite{goodfellow2014generative} model is to simultaneously find the parameters of a discriminator D that maximizes its classification accuracy, and the parameters of a generator G that maximizes the probability of D making a mistake. This objective corresponds to a min-max two-player game, which can be formulated as:
\begin{equation}
\begin{split}
    L(D, G) &=\underset{\theta_G}{min}\:\underset{\theta_D}{max}\:E_{x\sim p_{data}(x)}[logD(x)] \:\\
            &+E_{z\sim p_{z}(z)}[log(1-D(G(z)))],
\end{split}
\end{equation}
where $x$ is a real sample from an unknown distribution $P_{data}$, $z$ is the random noise, $L(D,G)$ represents the cost of training, and $\theta_G$ an $\theta_D$ are the parameters of G and D respectively.

How can we work out this concept in our work? As we mentioned earlier, the idea is to think of the baseline Mask R-CNN as a generator, and send its results to discriminators for validation. Therefore, our losses will be the combination of the generator's loss and the discriminators' losses. Formally, we write our losses as follows: 
\begin{equation}\label{eq:L_Generator}
L_{Generator} = L_{cls} + L_{bbox} + L_{mask} + L^{G_{b}}_{adv} +  L^{G_m}_{adv}, 
\end{equation}
where the first three terms in the right-hand side exactly define the loss of the baseline Mask R-CNN, and stand for loss of bounding box, class, and mask, respectively. The last two terms are our complementary losses where help us further optimize the generated results of the baseline. Thus,  $L^{G_{b}}_{adv}$ and  $L^{G_m}_{adv}$ stand for generator loss of the bounding box and the mask, respectively. 
To close the GANs formulation loop, we need a discriminator network to compete with $L^{G_{b}}_{adv}$, denote its loss $L^{D_{b}}_{adv}$, and one for  $L^{G_m}_{adv}$, denote its loss $L^{D_{m}}_{adv}$. Putting the discriminators' losses together we get:
\begin{equation}\label{eq:L_Discriminators}
L_{Discriminators} = L^{D_{b}}_{adv} +  L^{D_m}_{adv}.  
\end{equation}

Looking at Eq. (\ref{eq:L_Generator}) and (\ref{eq:L_Discriminators}), the only remaining terms to be defined are pairs of $(L^{G_{b}}_{adv}, L^{D_b}_{adv})$ and $(L^{G_{m}}_{adv}, L^{D_m}_{adv})$ which we will elaborate on per architecture design, in the following sections. 

\begin{figure}[!ht]\centering
\includegraphics[width=1\linewidth]{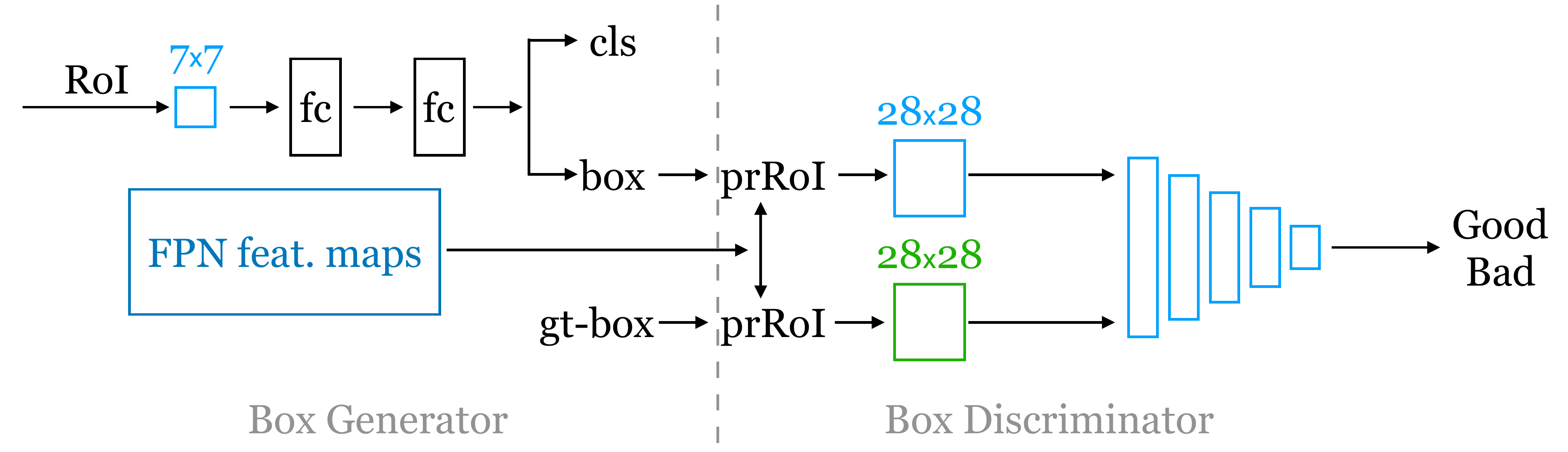} 
\caption{Box head architecture. We propose an adversarial training framework for object detection by extending the current box head architecture in Mask R-CNN with an FPN backbone \cite{he2017mask}. We design a new discriminator network, which takes in bounding box predictions from the box head and their ground truths, then extracts the corresponding regions of interest (RoI) from the FPN feature maps using Precise RoI Pooling (prRoI) \cite{jiang2018acquisition}. Finally, the discriminator outputs a score between $[0,1]$ that represents how good the bounding box prediction is. There are $5$ convolution layers in the discriminator network: each convolution layer is followed by a BatchNorm and a LeakyReLU layer. Note we only use convolution layers with the correct stride and kernel size to down-sample the bbox prediction from its original spatial dimension to our desired output shape.}
\label{fig:gan_box}
\end{figure}

\subsection{Network architecture}

\subsubsection{Box head:}
\textbf{Generator:} As shown in Fig.~\ref{fig:gan_box}, we utilize the box head of Mask R-CNN as our box generator; thus, instead of taking random noise as input like in a vanilla GAN, our generator now takes in RoI features, and outputs class and bounding box prediction. During training, we pick only the predicted, class-agnostic boxes and send them to the discriminator.

\textbf{Discriminator:} The purpose of the box discriminator is to look at the bounding box prediction from the generator, and tell whether they are good or bad. Practically, it is really difficult for any discriminator to evaluate bounding box predictions of objects simply by looking at the 4 coordinates. Our solution, therefore, is to actually send the feature map of the predicted boxes and its ground truth as fake and real samples to the discriminator. More importantly, in order to train the box head generator/discriminator with gradient descent, we need to enable backpropagation w.r.t \textit{bounding box coordinates}. For this purpose, we extract the region of interest using Precise RoI Pooling \cite{jiang2018acquisition} on the \textit{feature maps output} of FPN backbone \cite{he2017mask} for both box predictions and ground truths. This discriminator network consists of $5$ convolution layers followed by a BatchNorm and a LeakyReLU layer as shown in Fig.~\ref{fig:gan_box}. It takes in a multidimensional images of size $28\times28$ and outputs a score between $[0,1]$, which represents how good the prediction is (higher score is better). 

\textbf{Loss function:} As discussed in Eq. (\ref{eq:L_Generator}) and (\ref{eq:L_Discriminators}), we introduce an adversarial loss term to optimize our generator/discriminator framework as follows: 
\begin{equation}
L^{G_{b}}_{adv} = \frac{1}{N}\sum^N_{i=1}-log(D_{b}(G_{b}(RoI_i))), \\
\end{equation}
\begin{equation}
\begin{split}
    L^{D_{b}}_{adv} &= \frac{1}{N}\sum^N_{i=1}-(log(D_{b}(bb^{gt}_i)) \\
                    &+log(1-D_{b}(G_{b}(RoI_i)))),
\end{split}
\end{equation}
where $N$ is the mini-batch size, and $G_{b}(RoI_i)$ and $bb^{gt}_i$ denote the $i$-th bounding box prediction and its corresponding ground truth, respectively. Finally, $D_{b}(.)$ denotes the probability of an image being real. In a min-max setup, $L^{G_{b}}_{adv}$ encourages $G_{b}$ to generate a bounding box that can fool $D_{b}$, while $L^{D_{b}}_{adv}$ strengthens $D_{b}$'s ability to differentiate between real and fake bounding boxes.

\begin{figure}[!htb]\centering
\includegraphics[width=1\linewidth]{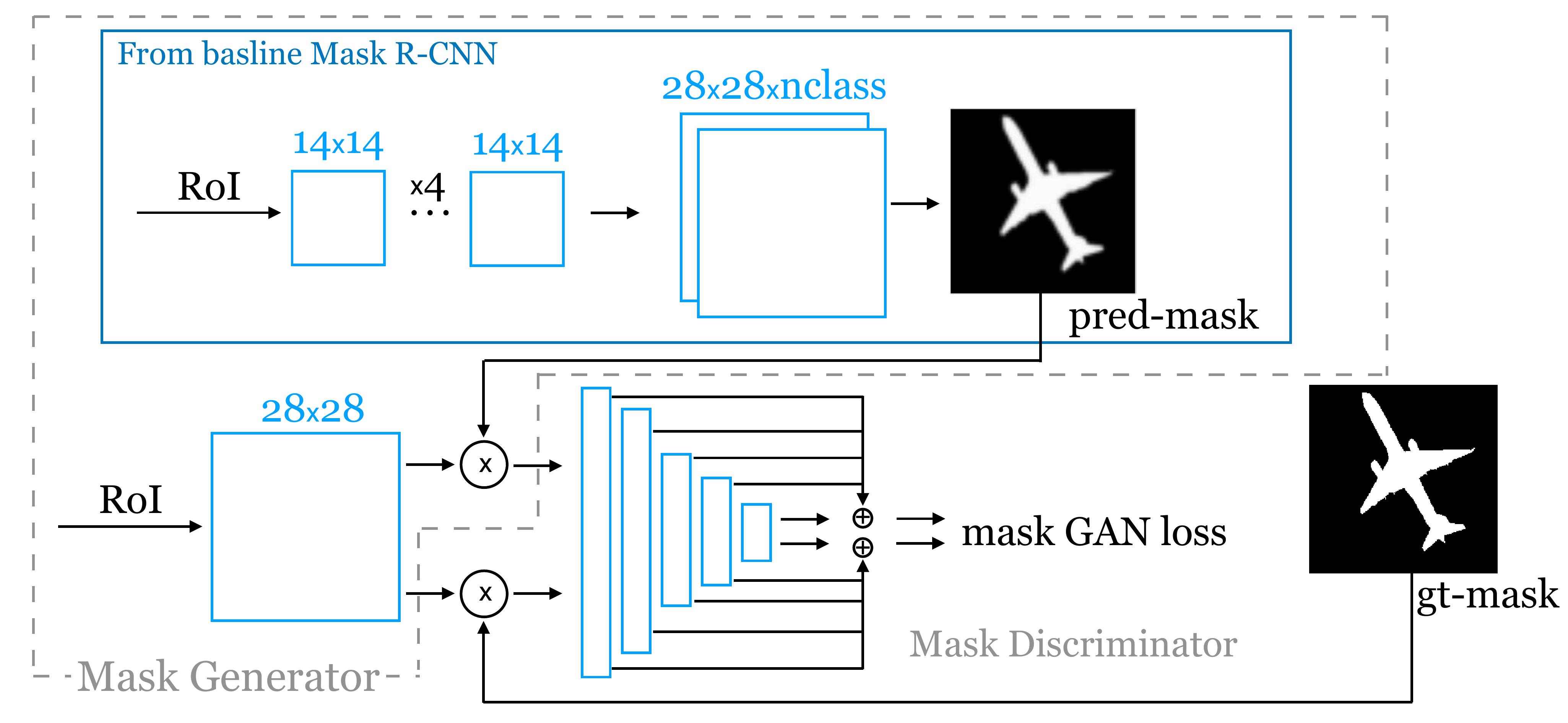} 
\caption{Mask head architecture. We utilize the mask head of Mask R-CNN \cite{he2017mask} as the generator, and build another discriminator network for the mask segmentation task. During training, we first pick the binary mask predictions of true objects (pred-mask) and their ground truths (gt-mask), calculate the pixel-wise multiplication between these masks and the corresponding regions of interest (RoI), and send the results to the discriminator as inputs. Outputs from each layer of the discriminator are flatten then concatenated and used for final GAN loss computation. Similar to the box head, the mask head also has 5 convolution layers in the discriminator network, where each convolution layer is followed by  a BatchNorm and a LeakyReLU layer.}
\label{fig:gan_mask}
\end{figure}

\subsubsection{Mask head:}

\textbf{Generator:} Same as the box head, we also adopt the mask head of Mask R-CNN as our mask generator. This network contains $4$ convolution layers followed by a convolution-transpose and an $1\times1$ convolution layer (Fig.~\ref{fig:gan_mask}). It takes RoI features as input, and output a binary mask prediction of size $28\times28$ for each objects class. During training, we extract the logits from the channel of predicted class, and send them to the discriminator.

\textbf{Discriminator:} As shown in Fig.~\ref{fig:gan_mask}, we feed two inputs to the mask discriminator, the binary mask prediction and its ground truth, both multiplied element-wise with the RoI features of size $28\times28$. This network has $5$ convolution layers and is constructed similar to an encoder such that the features' spatial resolution reduces as we go deeper. We concatenate these hierarchical features from all the layers into a single output and use it to compute the adversarial loss.

\textbf{Loss function:} Following the formulations in Eq. (\ref{eq:L_Generator}) and (\ref{eq:L_Discriminators}), the adversarial loss for optimizing the mask generator $L^{G_m}_{adv}$ and discriminator $L^{D_m}_{adv}$ can be formulated as $L^{D_m}_{adv}=-L^{G_m}_{adv}$, where: 
\begin{equation}\label{eq:head_loss}
L^{G_m}_{adv} = \frac{1}{N}\sum^N_{i=1}\norm{D_m(mask^{gt}_i) - D_m(G_m(RoI_i))}_1,
\end{equation}

where $N$ is the mini-batch size, and $G_m(RoI_i)$ and $mask^{gt}_i$ denote the $i$-th binary mask prediction and its corresponding ground truth, respectively. $L^{G_m}_{adv}$ encourages $G_m$ to minimize the differences between the mask prediction and ground truth based on the feedback from D, while  $L^{D_m}_{adv}$ guide $D_m$ to magnify those difference in reverse. We use $L_1$ perceptual distance as studied in \cite{isola2017image}.

\section{Experiments}
In this section, we present a thorough comparison between our model and the baseline, Mask R-CNN over several challenging public datasets, i.e., Phone Recycling dataset \cite{Jahanian_2019_CVPR_Workshops}, Cityscapes \cite{Cordts2016Cityscapes}, COCO \cite{lin2014microsoft}, and MICCAI 2015 Challenge \cite{DBLP:journals/corr/Sirinukunwattana16} on histology gland images. Ablative studies are provided in Supplementary materials and summarized here as well.

\subsection{Experimental settings and evaluation metrics}
For all experiments, we use the pretrained ResNet \cite{he2016deep} backbone with FPN architecture \cite{lin2017feature} to initialize our network. The detail implementation is based on the public Mask R-CNN benchmark project \cite{massa2018mrcnn} by Facebook on the Pytorch platform. We train the whole network with Stochastic Gradient Descent on two NVIDIA TITAN RTX (24GB memory each). 
More details on the training hyperparameters for each datasets will be mentioned in the following sections.

To have a fair comparison on the performance, we ensure that when training either vanilla baseline or GAN Mask R-CNN, they will both receive the same amount of training iterations. For COCO, all training schedules follow the benchmark scheme from official Facebook source code (Mask RCNN owner). Note that this training schedule from Facebook ensures that the model performance is similar regardless of numbers of GPUs. For the other datasets, to avoid undertraining, we train the model under various epoch settings and pick the best scheme with the highest validation accuracy for comparison.

During evaluation, we report the standard COCO metric: Average Precision (AP). AP is measured over multiple IoU's (Intersection of prediction and ground truth over Union of prediction and ground truth) thresholds ($[0.5:0.05:0.95]$); whereas $AP_{50}$, $AP_{75}$ are average precision at IoU threshold of $0.5$ and $0.75$ respectively. , $AP_S, AP_M, AP_L$ are average precision computed at different object scales. Scale is determined by the number of pixels in the segmentation masks: small objects ($area < 32^2$ pixels), medium objects ($32^2 < area < 96^2$), and large objects ($area > 96^2$).

\subsection{Training GAN Mask R-CNN}
We train our network by using backpropagation from original Mask R-CNN losses and the new adversarial losses. The generator and the discriminator are trained in an alternating fashion. We first freeze the generator and train the discriminator. As discriminator training tries to figure out how to distinguish real data from fake, it has to learn how to recognize the generator's flaws. Similarly, we freeze the discriminator when training the generator. As training progresses, both the generator and discriminator networks become more powerful, and eventually, the generator will be able to produce predictions that are very close to the ground truths. 

In all experiments, we used the SGD optimizer with weight decay $0.0001$ and momentum $0.9$. Additionally, we used the same learning rate schedule for both the generator and the discriminator except that the discriminator's learning rate magnitude is one fifth of the generator's one.

\begin{figure*}[!htb]\centering
\includegraphics[width=1\linewidth]{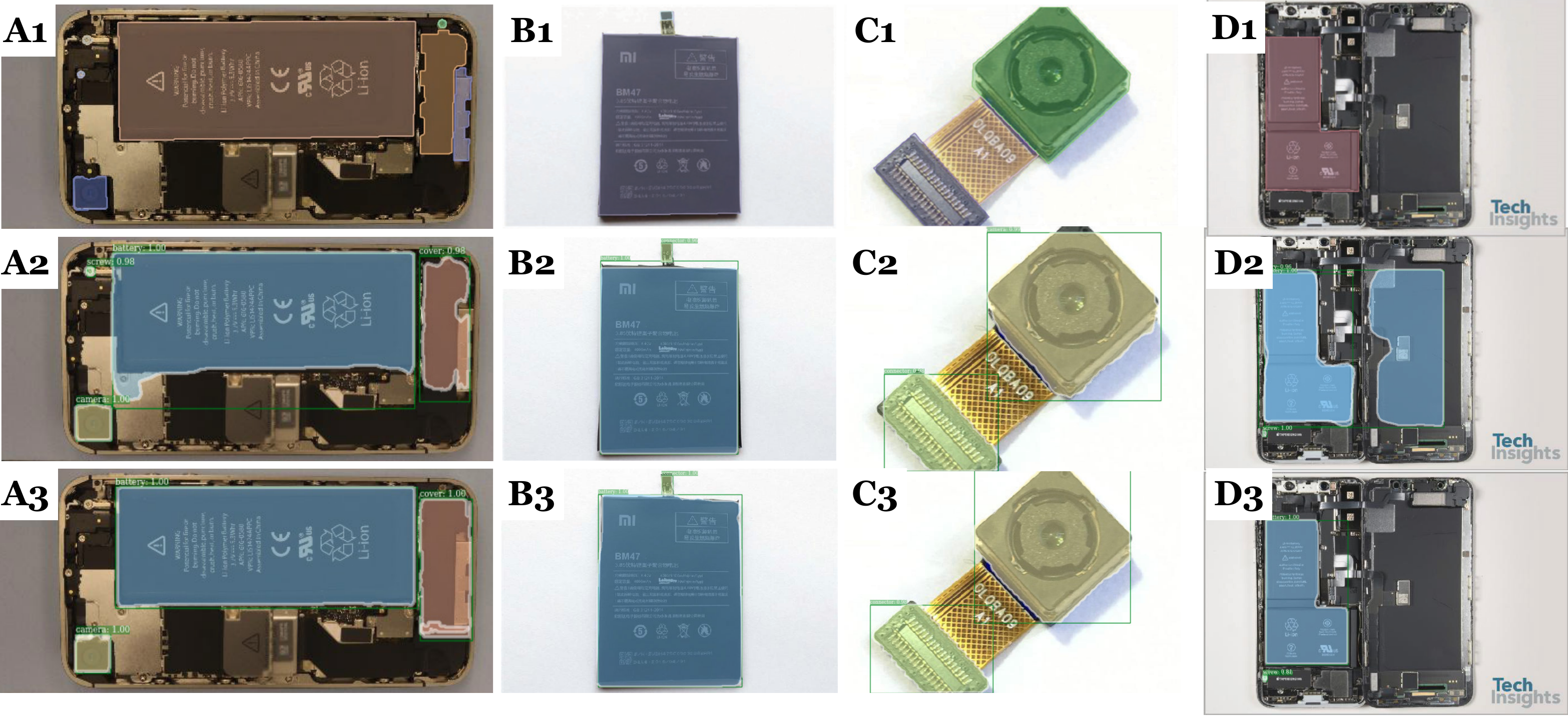} 
\caption{Instance Segmentation Results on the Phone Recycling \textit{validation} set: Mask R-CNN (second row) vs our model (third row). The images' ground truths are showed on the first row for reference. Overall, our model produces more precise bounding box detection (i.e. box predictions in B2 vs B3 and C2 vs C3) and  and sharper segmentation especially around the objects' boundaries (i.e. battery segmentation in A2 vs A3 and D2 vs D3).}
\label{fig:phone_results}
\end{figure*}

\subsection{Phone recycling dataset}
The Phone Recycling dataset introduced in \cite{Jahanian_2019_CVPR_Workshops} is used for high-precision disassembling task. It consists of $533$ high-resolution images of cellphone individual components as well as its different layers. There are $10$ cellphone models used in the dataset: \emph{Apple iPhone 3GS}, \emph{iPhone 4}, \emph{iPhone 4S}, \emph{iPhone 6}, \emph{Samsung GT-i8268 Galaxy}, \emph{Samsung S4 Active}, \emph{Samsung Galaxy S6} and \emph{S6 Edge}, \emph{Samsung S8 Plus}, \emph{Pixel 2 XL}, \emph{Xiomi Note}, \emph{HTC One}, \emph{Huawei Mate 8, 9, 10}, and  \emph{P8 Lite}, and $11$ object categories in the instance segmentation task.\\

\noindent\textbf{Implementation:} We first split the dataset into $90\%$ for training and $10\%$ for validation, and then apply augmentation on the training set. By adding random noise, Gaussian blur, sharpness, or changing the lightness, as well as constant normalization, we produce more than $4000$ training images. For validation, we pick at least one image from each cellphone model which is not shown to our network during training.

We use ResNet-101-FPN as the backbone for both the baseline Mask R-CNN and our network. To reduce overfitting, we train with the image scale (shorter side) randomly sampled from $[400, 600]$, and inference at 400 pixels. We use batch size of $8$ images per GPU and train the network for $27k$ iterations. We start with a learning rate of $0.01$ and reduce it to $0.001$ and $0.0001$ at $18k$ and $24k$ iteration respectively.

\begin{table}[!htb]
\centering
\caption{Phone recycling object detection (A) and segmentation (B) results (bounding box and mask AP). Comparisons between the baseline Mask-RCNN and our model on the Phone Recycling \textit{validation} set. ResNet-101-FPN is used in both cases. Our model using GAN framework achieves $3.6\%$ and 2.1$\%$ improvement in bounding box and mask AP respectively compared to Mask-RCNN.}
\begin{tabular}{c|c|ccc|ccc}
                   &       & $AP$            & $AP_{50}$       & $AP_{75}$       & $AP_{S}$        & $AP_{M}$        & $AP_{L}$        \\ \hline
\multirow{2}{*}{A} & MRCNN & 0.5486          & 0.7333          & 0.6261          & 0.3327          & 0.5071          & 0.7462          \\
                   & Ours  & \textbf{0.5844} & \textbf{0.7746} & \textbf{0.6448} & \textbf{0.3833} & \textbf{0.5480} & \textbf{0.7586} \\ \hline
\multirow{2}{*}{B} & MRCNN & 0.5718          & 0.7242          & 0.6439          & 0.3195          & 0.5247          & 0.7735          \\
                   & Ours  & \textbf{0.5923} & \textbf{0.7607} & \textbf{0.6525} & \textbf{0.3615} & \textbf{0.5549} & \textbf{0.7846}
\end{tabular}
\label{tab:phone_acc}
\end{table}

\noindent\textbf{Result:} Table \ref{tab:phone_acc} show the comparison of our network with baseline Mask R-CNN for both object detection and mask segmentation tasks on the phone \textit{validation} set. We observe that our proposed approach outperforms the baseline in both tasks, with an average improvement in AP metric of $3\%$ on bounding box detection ($58\%$ vs $55\%$) and $2\%$ on mask segmentation ($59\%$ vs $57\%$). 

One of the main challenges of this dataset is that there are multiple small components with only a few pixels gap between them; most of the components are flat and usually overlap each other (to save space during manufacturing). In general, it is difficult to identify small object due to the compactness and occlusion \cite{Jahanian_2019_CVPR_Workshops}. Our method shows noticeable improvement on this category, achieving around $15\%$ improvement on both bounding box detection (from $33\%$ to $38\%$) and mask segmentation (from $32\%$ to $36\%$). This is because of using GANs where the framework use more structure than individual pixels for learning.

Figure~\ref{fig:phone_results} presents some qualitative results of our method and the baseline Mask R-CNN with the corresponding ground truth. We observe that our method not only achieve better quantitative result but also have better edge prediction compared to Mask R-CNN (i.e., battery segmentation in A3 vs A2 and D3 vs D2 in Fig.~\ref{fig:phone_results} as well as falsely segmenting the cover as an extra battery in D2). while finding perceptual metrics that correctly and reliably tells us about boundaries and junctions of edges is an open area of research, we can qualitatively examine the results here. For disassembling tasks on small circuit boards like cellphone, finding crisp boundaries (i.e., finding a clear gap between two components for positioning a tool for prying/pulling out the component) is critical. Therefore, it is clear that our method provides more reliable detection than the baseline.

\begin{figure*}[!htb]\centering
\includegraphics[width=0.95\textwidth]{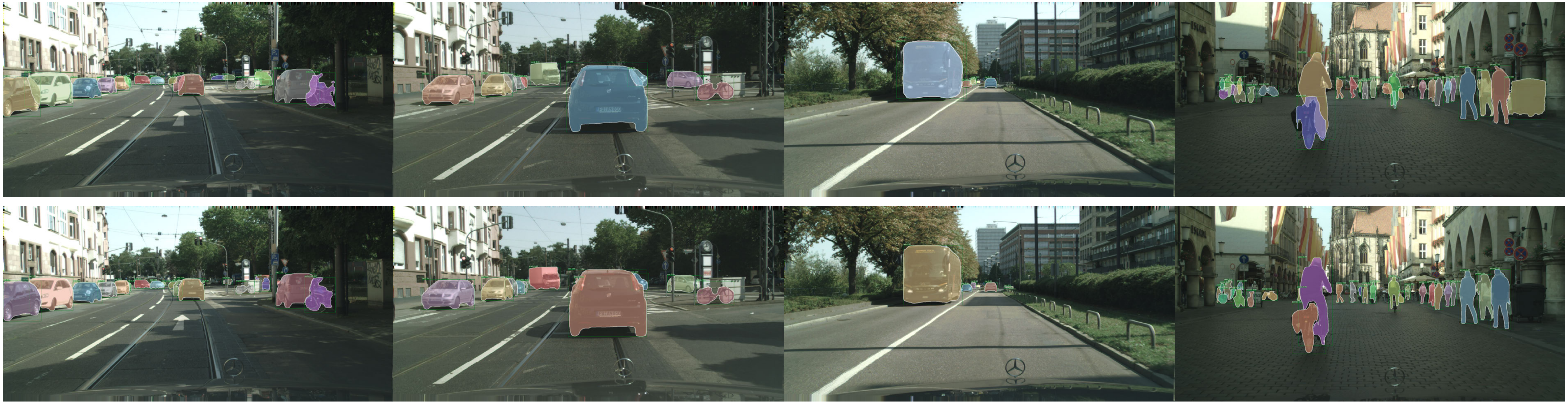} 
\caption{Instance segmentation results on the Cityscapes \textit{validation} set: Mask R-CNN (first row) vs our model (second row). Comparing the two result, we observe that our model produces sharper object boundary predictions, especially in where there are multiple of clusters of objects that overlap each other.}
\label{fig:city_results}
\end{figure*}

\begin{table*}[]
\centering
\scriptsize
\caption{Cityscape dataset object detection (A) and segmentation (B) results (bounding box and mask AP). Comparisons between the baseline Mask-RCNN and our model on the Cityscapes \textit{validation} set. ResNet-50-FPN is used in both cases. On average, our model achieves $1.7\%$ and 2.3$\%$ improvement in bounding box and mask AP metric respectively compared to Mask-RCNN.}
\resizebox{\textwidth}{!}{%
\begin{tabular}{c|c|c|c|c|c|c|c|c|c|c|c|c|c}
 &
   &
  Metric &
  Average &
  Person &
  Rider &
  Car &
  Truck &
  Bus &
  Caravan &
  Trailer &
  Train &
  Motocycle &
  Bicycle \\ \hline
\multirow{6}{*}{A} &
  \multirow{3}{*}{MRCNN} &
  $AP$ &
  0.406 &
  0.378 &
  0.459 &
  0.545 &
  0.385 &
  \textbf{0.572} &
  0.499 &
  0.326 &
  0.268 &
  \textbf{0.32} &
  0.313 \\
 &
   &
  $AP_{50}$ &
  0.73 &
  0.721 &
  0.806 &
  0.84 &
  0.655 &
  \textbf{0.827} &
  0.839 &
  0.641 &
  \textbf{0.661} &
  \textbf{0.682} &
  0.629 \\
 &
   &
  $AP_{75}$ &
  0.373 &
  0.349 &
  0.457 &
  0.575 &
  0.379 &
  \textbf{0.679} &
  0.484 &
  0.17 &
  \textbf{0.155} &
  0.208 &
  0.273 \\ \cline{2-14} 
 &
  \multirow{3}{*}{Ours} &
  $AP$ &
  \textbf{0.423} &
  \textbf{0.383} &
  \textbf{0.482} &
  \textbf{0.552} &
  \textbf{0.403} &
  0.562 &
  \textbf{0.592} &
  \textbf{0.359} &
  \textbf{0.27} &
  0.309 &
  \textbf{0.317} \\
 &
   &
  $AP_{50}$ &
  \textbf{0.745} &
  \textbf{0.723} &
  \textbf{0.821} &
  \textbf{0.844} &
  \textbf{0.704} &
  0.787 &
  \textbf{0.888} &
  \textbf{0.764} &
  0.632 &
  0.646 &
  \textbf{0.638} \\
 &
   &
  $AP_{75}$ &
  \textbf{0.409} &
  \textbf{0.364} &
  \textbf{0.497} &
  \textbf{0.58} &
  \textbf{0.394} &
  0.638 &
  \textbf{0.708} &
  \textbf{0.298} &
  0.112 &
  \textbf{0.221} &
  \textbf{0.276} \\ \hline
\multirow{6}{*}{B} &
  \multirow{3}{*}{MRCNN} &
  $AP$ &
  0.403 &
  0.329 &
  0.334 &
  0.535 &
  0.391 &
  0.594 &
  0.557 &
  0.353 &
  \textbf{0.442} &
  \textbf{0.271} &
  0.219 \\
 &
   &
  $AP_{50}$ &
  0.724 &
  0.693 &
  0.811 &
  0.819 &
  0.624 &
  \textbf{0.817} &
  \textbf{0.957} &
  0.642 &
  0.669 &
  \textbf{0.652} &
  0.561 \\
 &
   &
  $AP_{75}$ &
  0.365 &
  0.28 &
  0.148 &
  0.561 &
  0.417 &
  \textbf{0.696} &
  0.484 &
  0.288 &
  \textbf{0.507} &
  \textbf{0.145} &
  0.121 \\ \cline{2-14} 
 &
  \multirow{3}{*}{Ours} &
  $AP$ &
  \textbf{0.416} &
  \textbf{0.34} &
  \textbf{0.353} &
  \textbf{0.545} &
  \textbf{0.414} &
  \textbf{0.598} &
  \textbf{0.597} &
  \textbf{0.407} &
  0.425 &
  0.258 &
  \textbf{0.228} \\
 &
   &
  $AP_{50}$ &
  \textbf{0.736} &
  \textbf{0.703} &
  \textbf{0.844} &
  \textbf{0.825} &
  \textbf{0.676} &
  0.792 &
  0.888 &
  \textbf{0.763} &
  \textbf{0.701} &
  0.59 &
  \textbf{0.582} \\
 &
   &
  $AP_{75}$ &
  \textbf{0.398} &
  \textbf{0.288} &
  \textbf{0.192} &
  \textbf{0.576} &
  \textbf{0.448} &
  0.674 &
  \textbf{0.595} &
  \textbf{0.483} &
  0.448 &
  0.143 &
  \textbf{0.136}
\end{tabular}%
}
\label{tab:city_acc}
\end{table*}

\subsection{Cityscapes dataset}
The Cityscapes dataset \cite{Cordts2016Cityscapes} focuses on semantic understanding of urban street scenes. It contains high-resolution images ($2048\times1024$ pixels) from over $50$ cities, with a time span of several months under good/medium weather conditions. This dataset has $5k$ annotated images with fine annotations ($2975$ train, $500$ val and $1525$ test images) and $20k$ annotated images with coarse annotations. For instance segmentation task, we use only the fine annotated images with $10$ object categories involved to train our network. \\

\noindent\textbf{Implementation:} We use ResNet-50-FPN as the backbone for both the baseline Mask R-CNN and our network. To reduce overfitting, we train with the image scale (shorter side) randomly sampled from $[800, 1024]$, and inference at $1024$ pixels. We use batch size of $4$ images per GPU and train the network for $24k$ iterations. We start with a learning rate of $0.01$ and reduce it to $0.001$ at $18k$ iteration.

\noindent\textbf{Result:} We report instance segmentation results using the evaluation helper tool provided along with the Cityscapes dataset. Table \ref{tab:city_acc} summarize the detail comparison between our network and Mask R-CNN in both object detection and mask segmentation tasks on the \textit{validation} set. On average, our method achieves around $2-3\%$ increase in all AP thresholds on both tasks; i.e. for object detection, AP: from $40.6\%$ to $42.3\%$, $AP_{50}$: from $73\%$ to $74.5\%$, and $AP_{75}$: from $37.3\%$ to $40.9\%$, and for mask segmentation, AP: from $40.3\%$ to $41.\%$, $AP_{50}$: from $72.4\%$ to $73.6\%$, and $AP_{75}$: from $36.5\%$ to $39.8\%$. For further analysis, we also report AP metrics for all $10$ object categories. We observe that our method gains consistent performance improvement for most classes while maintaining comparable performance for other object classes. 

More qualitative results of our network and Mask R-CNN on Cityscapes are shown in Fig.~\ref{fig:city_results}. 

\subsection{COCO dataset}
The MS COCO \cite{lin2014microsoft} is a large-scale, richly annotated dataset for object detection and segmentation. It comprises of images depicting complex everyday scenes of common objects in their natural context. In this experiment, we train our network on the 2017 COCO dataset for instance segmentation, which contains $118k$ train and $5k$ validation images of $80$ object categories. 

\begin{figure*}[!htb]\centering
\includegraphics[width=0.95\textwidth]{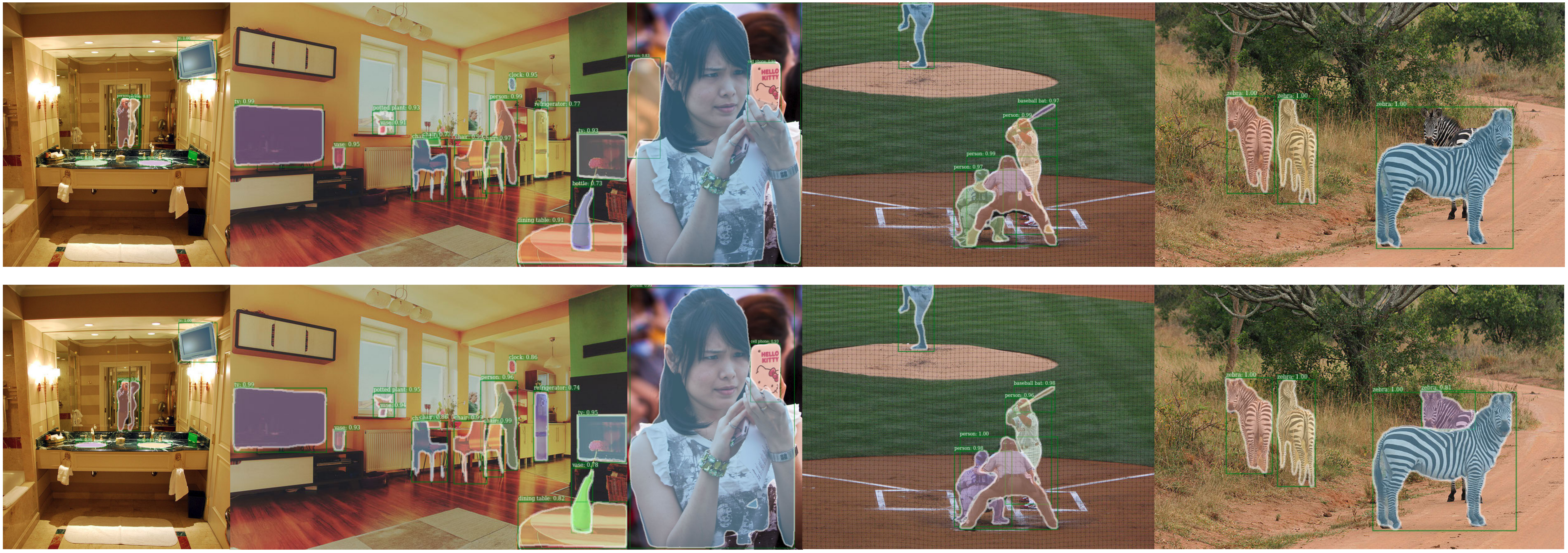} 
\caption{Instance segmentation results on the 2017 COCO \textit{validation} set. First row: baseline Mask R-CNN results. Second row: results of our model. From left to right, we observe that our model produces both better boundaries between different instances in a clutter (human mask in the fourth image) and more correct predictions (in the first image of human in front of mirror, and missing zebra in the fifth image.)}
\label{fig:coco_results}
\end{figure*}

\begin{table}[!htb]
\centering
\caption{The 2017 COCO object detection (A) and segmentation (B) results (bounding box and mask AP). Comparisons between the baseline Mask-RCNN and our model on the \textit{validation} set. ResNet-101-FPN is used in both cases. Our model using GAN framework achieves comparable performance in bounding box AP and 0.8$\%$ improvement in mask AP metric compared to Mask-RCNN.}
\begin{tabular}{c|c|ccc|ccc}
                   &       & $AP$            & $AP_{50}$       & $AP_{75}$       & $AP_{S}$        & $AP_{M}$        & $AP_{L}$        \\ \hline
\multirow{2}{*}{A} & MRCNN & 0.3992          & 0.6116          & \textbf{0.4387} & \textbf{0.2336} & 0.4328          & 0.5183          \\
                   & Ours  & \textbf{0.3994} & \textbf{0.6143} & 0.4342          & 0.2317          & \textbf{0.4332} & \textbf{0.5269} \\ \hline
\multirow{2}{*}{B} & MRCNN & 0.3604          & 0.5828          & 0.3835          & 0.1606          & 0.3892          & 0.5299          \\
 & Ours & \textbf{0.3681} & \textbf{0.5855} & \textbf{0.3919} & \textbf{0.1685} & \textbf{0.3988} & \textbf{0.5452}
\end{tabular}%
\label{tab:coco_acc}
\end{table}

\noindent\textbf{Implementation:} We use ResNet-101-FPN as the backbone for both the baseline Mask R-CNN and our network.  Images are resized such that their scale (shorter edge) is $800$ pixels. We use batch size of $4$ images per GPU and train the network for $180k$ iterations. We start with a learning rate of $0.01$ and reduce it to $0.001$ and $0.0001$ at $120k$ and $160k$ iteration respectively. Note that this training settings is equivalent to the settings used in the original Mask R-CNN paper \cite{he2017mask} according to scheduling rules from Facebook Detectron \cite{Detectron2018}.

\noindent\textbf{Result:} Table \ref{tab:coco_acc} shows the comparison of our network with Mask R-CNN for mask segmentation task on the 2017 COCO \textit{validation} set. We observe that our proposed approach outperforms the baseline Mask R-CNN, achieving $0.8\%$ improvement in AP metric. 

Qualitative results of our network and Mask R-CNN on the 2017 COCO \textit{validation} set are shown in Fig.~\ref{fig:coco_results}.

\subsection{Gland segmentation dataset}
This dataset is provided by the MICCAI 2015 Challenge Contest \cite{DBLP:journals/corr/Sirinukunwattana16} on gland segmentation in histology images. There are $165$ labeled images of Hematoxylin and Eosin (H$\&$E) stained slides, consisting of a variety of histologic grades. Original images are in different sizes, but most of them are $775\times522$ pixels. The training set has $85$ images, in which there are $37$ benign sections and $48$ malignant ones; the \textit{test} set has $80$ images, in which there are $37$ benign sections and $43$ malignant one.

\begin{figure*}[!htb]\centering
\includegraphics[width=0.95\textwidth]{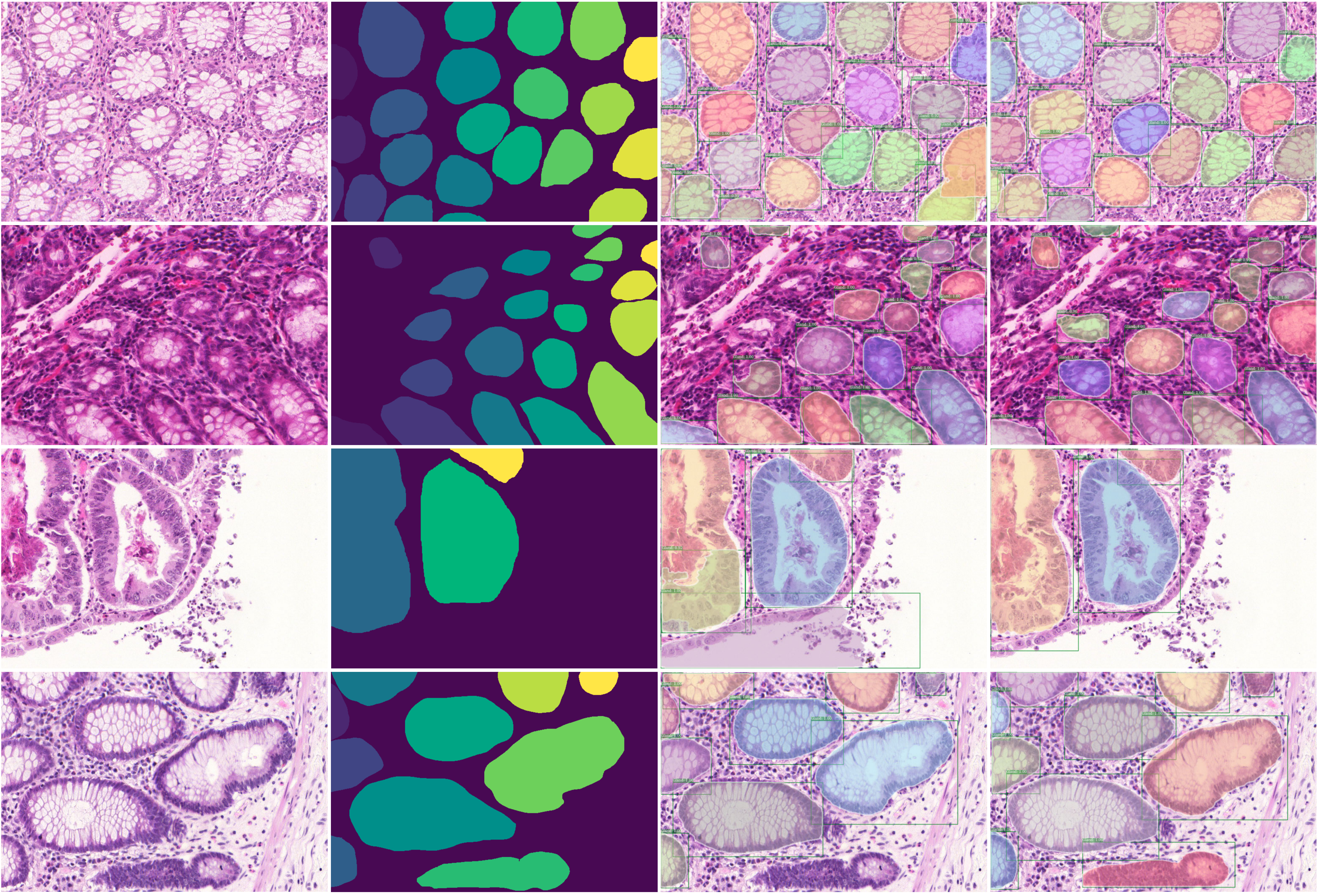} 
\caption{Instance segmentation results on the gland segmentation \textit{test} set. First column: image, second column: annotations, third column: results of baseline Mask R-CNN, fourth column: results of our model. We observe that our model produces more accurate gland detection and better boundary detection between different instances of glands.}
\label{fig:medical_results}
\end{figure*}

\noindent\textbf{Implementation:} We use ResNet-50-FPN as the backbone for both the baseline Mask R-CNN and our network. To reduce overfitting, we train with the image scale (shorter side) randomly sampled from $[400, 522]$, and inference at 522 pixels. We use batch size of $4$ images per GPU and train the network for $2k$ iterations. We start with a learning rate of $0.01$ and reduce it to $0.001$ at $1.5k$ iterations.

\begin{table}[!htb]
\centering
\caption{The MICCAI 2015 medical gland (A) and segmentation (B) results (bounding box and mask AP). Comparisons between the baseline Mask-RCNN and our model on the Gland \textit{test} set. ResNet-50-FPN is used in both cases. Our model using GAN framework achieves $1.7\%$ and 2.7$\%$ improvement in bounding box and mask AP metric respectively compared to Mask-RCNN.}
\begin{tabular}{c|c|ccc|cc}
                   &       & $AP$            & $AP_{50}$       & $AP_{75}$       & $AP_{M}$        & $AP_{L}$        \\ \hline
\multirow{2}{*}{A} & MRCNN & 0.5689          & 0.8174          & 0.6419          & 0.5940          & 0.5717          \\
                   & Ours  & \textbf{0.5855} & \textbf{0.8398} & \textbf{0.6668} & \textbf{0.6057} & \textbf{0.5983} \\ \hline
\multirow{2}{*}{B} & MRCNN & 0.5892          & 0.8014          & 0.9814          & 0.589           & 0.6189          \\
                   & Ours  & \textbf{0.6156} & \textbf{0.8236} & \textbf{0.7054} & \textbf{0.6054} & \textbf{0.6516}
\end{tabular}%
\label{tab:gland_acc}
\end{table}
\noindent\textbf{Result:} We compare the performance of Mask R-CNN and our model on the Gland Segmentation Challenge as shown on Table \ref{tab:gland_acc}. Note that since there is no small objects ($area < 32^2$ pixels) in the \textit{test} set, we will not report on $AP_S$ metric. We observe from the tables that our model outperforms the baseline Mask R-CNN in both tasks, with an average improvement in AP metric of $2\%$ on bounding box detection ($57\%$ to $59\%$) and $3\%$ on mask segmentation ($59\%$ to $62\%$).

Qualitative results of our network and Mask R-CNN on the Gland Segmentation are shown in Fig.~\ref{fig:medical_results}. 
The excessive background noise, the heterogeneous shapes of objects, the small gap between objects and coalescing pixels are known challenges of this dataset \cite{xu2017gland}. We observe that our results much better account for these cases when comparing to the baseline ones.


\section{Conclusions}

We redefined mask prediction as mask generation in GANs and introduced designs on the general segmentation framework in Mask R-CNN for effective GANs training. 
This design obtains a loss that is learned as oppose to a hand-engineered one, and thus makes the framework more generalizable for learning domain-agnostic datasets.  
We show this GANs framework quantitatively and qualitatively performs better than the original Mask R-CNN. For our experiments, we chose four datasets in various domains from regular shapes to heterogeneous and coalescing shapes, and simply trained our model on them, without any extra settings. 
We show the results give crispier boundaries, and better accounts for seeing structures and geometries as well as small details.



{
{\small
\bibliographystyle{ieee_fullname}
\bibliography{egbib}
}
}

\section{Appendix: Ablation Studies}

We performed some ablation studies, for number of the layers in Discriminator, and for isolating the the adversarial loss. The results indicate increasing number of the layers from three to five can slightly increase the performance. On the other hand, training with adversarial terms alone gives comparable performance as in the baseline while combining the two loss terms achieve the best overall performance. More details on these experiments will be provided in the supplementary materials.

\subsubsection{Analysis of the Objective Function}
How important are the adversarial components in the objective function of GAN Mask R-CNN? Can we train the framework using the adversarial components only?. To answer these questions, we isolate the effect of the adversarial terms by removing the second and third term of $L_{Generator}$ in Eqn. \ref{eq:L_Generator} and retrain the GAN Mask R-CNN with this new objective function. 

\begin{figure}[!ht]\centering
\includegraphics[width=\textwidth]{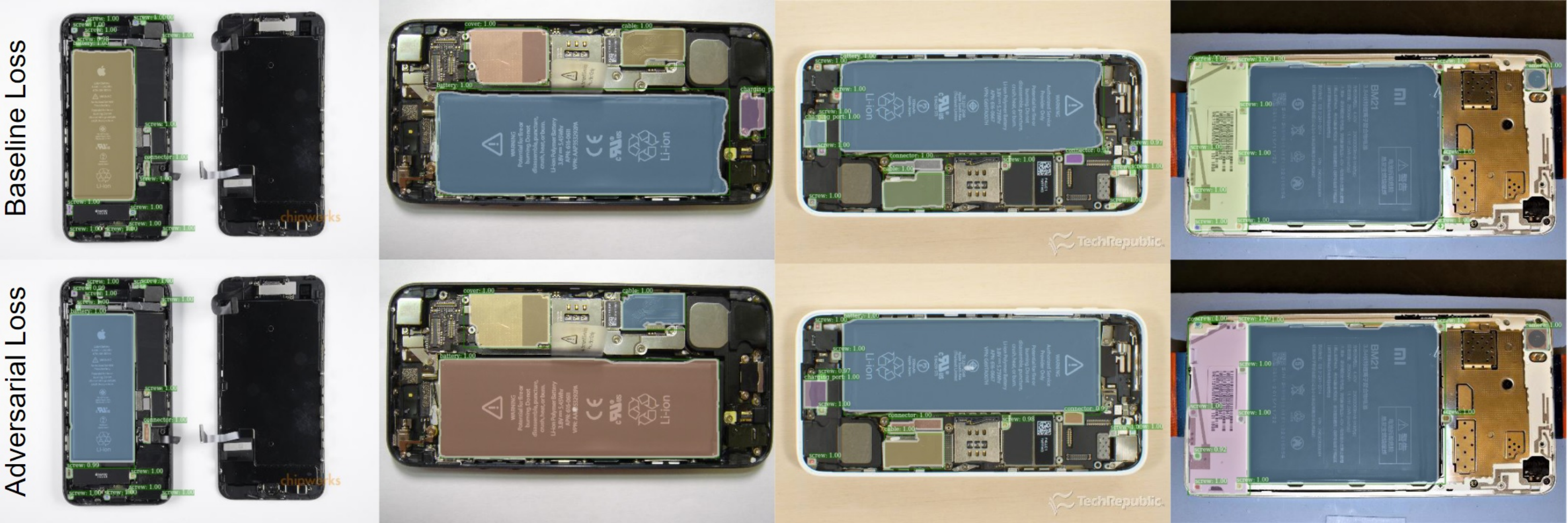} 
\caption{Instance segmentation result on different objective function. Each row shows results trained under a different loss.}
\label{fig:compareloss}
\end{figure}

Table \ref{tab:LossComapre} and Fig.~\ref{fig:compareloss} show the quantitative and qualitative result of our experiment. From table \ref{tab:LossComapre}, we observe that training our framework with adversarial terms alone still gives comparable performance as in the baseline. Furthermore, we can clearly see the effect of adversarial term in pushing the model to produce better edge detection as shown in Fig.~\ref{fig:compareloss}.

\begin{table*}[!htb]
\centering
\caption{AP metric score for different objective function, evaluated on the Phone Recycling validation set. We observe that using the our combined loss give the best performance.}
\begin{tabular}{c|c|c}
Loss Function               & $AP_{mask}$    & $AP_{bbox}$    \\ \hline
Baseline Loss               & \textbf{0.5718}          & 0.5486          \\
Adversarial Loss            & 0.5684          & \textbf{0.5620}          
\end{tabular}
\label{tab:LossComapre}
\end{table*}

\subsubsection{Analysis of the Discriminator architecture}
One of the main challenge in training GAN is that we need to make the generator and the discriminator equally powerful so they can learn effectively thorough competition. In this section, we run ablation studies on the number of convolution layers used in the discriminator. Given the 28x28 size of the input to the Discriminator, we tried three different design choices for the number of convolution layers. Table \ref{tab:NumDLayers} compares the AP metric score for both mask segmentation and object detection on Phone Recycling validation set. The table shows that using a shallow discriminator architecture may hurt the overall performance while using a deeper one can achieve superior result.

\begin{table*}[!htb]
\centering
\caption{AP metric score for different Discriminator architecture, evaluated on the Phone Recycling validation set. We observe that using 5 conv layers in Discriminator architecture give the best performance.}
\begin{tabular}{c|c|c}
\begin{tabular}[c]{@{}c@{}}Number of Conv layers\\ in Discriminator\end{tabular} & $AP_{mask}$    & $AP_{bbox}$    \\ \hline
3                                                                                & 0.5624          & 0.547           \\
4                                                                                & 0.574           & 0.55            \\
5                                                                                & \textbf{0.5923} & \textbf{0.5844}
\end{tabular}
\label{tab:NumDLayers}
\end{table*}

\end{document}